\documentclass[conference,compsoc]{IEEEtran}

\usepackage[nocompress]{cite}

\usepackage[pdftex]{graphicx}

\usepackage{amsmath}
\usepackage{array}
\begin{document}
%
\title{Improving the Performance of Neural Networks in Regression Tasks\\Using \emph{Drawering}}

\author{
\IEEEauthorblockN{Konrad \.Zo\l{}na}
\IEEEauthorblockA{Jagiellonian University, Cracow, Poland\\
RTB House, Warsaw, Poland\\
Email: konrad.zolna@im.uj.edu.pl, konrad.zolna@rtbhouse.com}}

\maketitle

\begin{abstract}
The method presented extends a given regression neural network to make its performance improve. The modification affects the learning procedure only, hence the extension may be easily omitted during evaluation without any change in prediction. It means that the modified model may be evaluated as quickly as the original one but tends to perform better.

This improvement is possible because the modification gives better expressive power, provides better behaved gradients and works as a regularization. The knowledge gained by the temporarily extended neural network is contained in the parameters shared with the original neural network.

The only cost is an increase in learning time.
\end{abstract}


%
\IEEEpeerreviewmaketitle

\section{Introduction}
Neural networks and especially deep learning architectures have become more and more popular recently \cite{lecun2015deep}. We believe that deep neural networks are the most powerful tools in a majority of classification problems (as in the case of image classification \cite{resnet}). Unfortunately, the use of neural networks in regression tasks is limited and it has been recently showed that a softmax distribution of clustered values tends to work better, even when the target is continuous \cite{wavenet}. In some cases seemingly continuous values may be understood as categorical ones (e.g. image pixel intensities) and the transformation between the types is straightforward \cite{Oord16}. However, sometimes this transformation cannot be simply incorporated (as in the case when targets span a huge set of possible values). Furthermore, forcing a neural network to predict multiple targets instead of just a single one makes the evaluation slower.

We want to present a method which fulfils the following requirements:
\begin{itemize}
\item gains advantage from the categorical distribution which makes a prediction more accurate,
\item outputs a single value which is a solution to a given regression task,
\item may be evaluated as quickly as in the case of the original regression neural network.
\end{itemize}

The method proposed, called \emph{drawering}, bases on temporarily extending a given neural network that solves a regression task. That modified neural network has properties which improve learning. Once training is done, the original neural network is used standalone. The knowledge from the extended neural network seems to be transferred and the original neural network achieves better results on the regression task.

The method presented is general and may be used to enhance any given neural network which is trained to solve any regression task. It also affects only the learning procedure.

\section{Main idea}
\subsection{Assumptions}
The method presented may be applied for a regression task, hence we assume:
\begin{itemize}
\item the data $D$ consists of pairs $(x_i,y_i)$ where the input $x_i$ is a fixed size real valued vector and target $y_i$ has a continuous value,
\item the neural network architecture $f(\cdot)$ is trained to find a relation between input $x$ and target $y$, for \mbox{$(x,y) \in D$},
\item a loss function $\mathcal{L}_f$ is used to asses performance of $f(\cdot)$ by scoring $\sum_{(x,y)\in D} \mathcal{L}_f(f(x), y)$, the lower the better.
\end{itemize}

\subsection{Neural network modification} \label{firtsMentionOfPercentiles}
In this setup, any given neural network $f(\cdot)$ may be understood as a composition $f(\cdot) = g(h(\cdot))$, where $g(\cdot)$ is the last part of the neural network $f(\cdot)$ i.e. $g(\cdot)$ applies one matrix multiplication and optionally a non-linearity. In other words, a vector $z = h(x)$ is the value of last hidden layer for an input $x$ and the value $g(z)$ may be written as $g(z) = \sigma (Gh(x))$ for a matrix $G$ and some function $\sigma$ (one can notice that $G$ is just a vector). The job which is done by $g(\cdot)$ is just to squeeze all information from the last hidden layer into one value.

In simple words, the neural network $f(\cdot)$ may be divided in two parts, the first, core $h(\cdot)$, which performs majority of calculations and the second, tiny one $g(\cdot)$ which calculates a single value, a prediction, based on the output of $h(\cdot)$.

\begin{figure}[!t]
\centering
\includegraphics[width=\columnwidth]{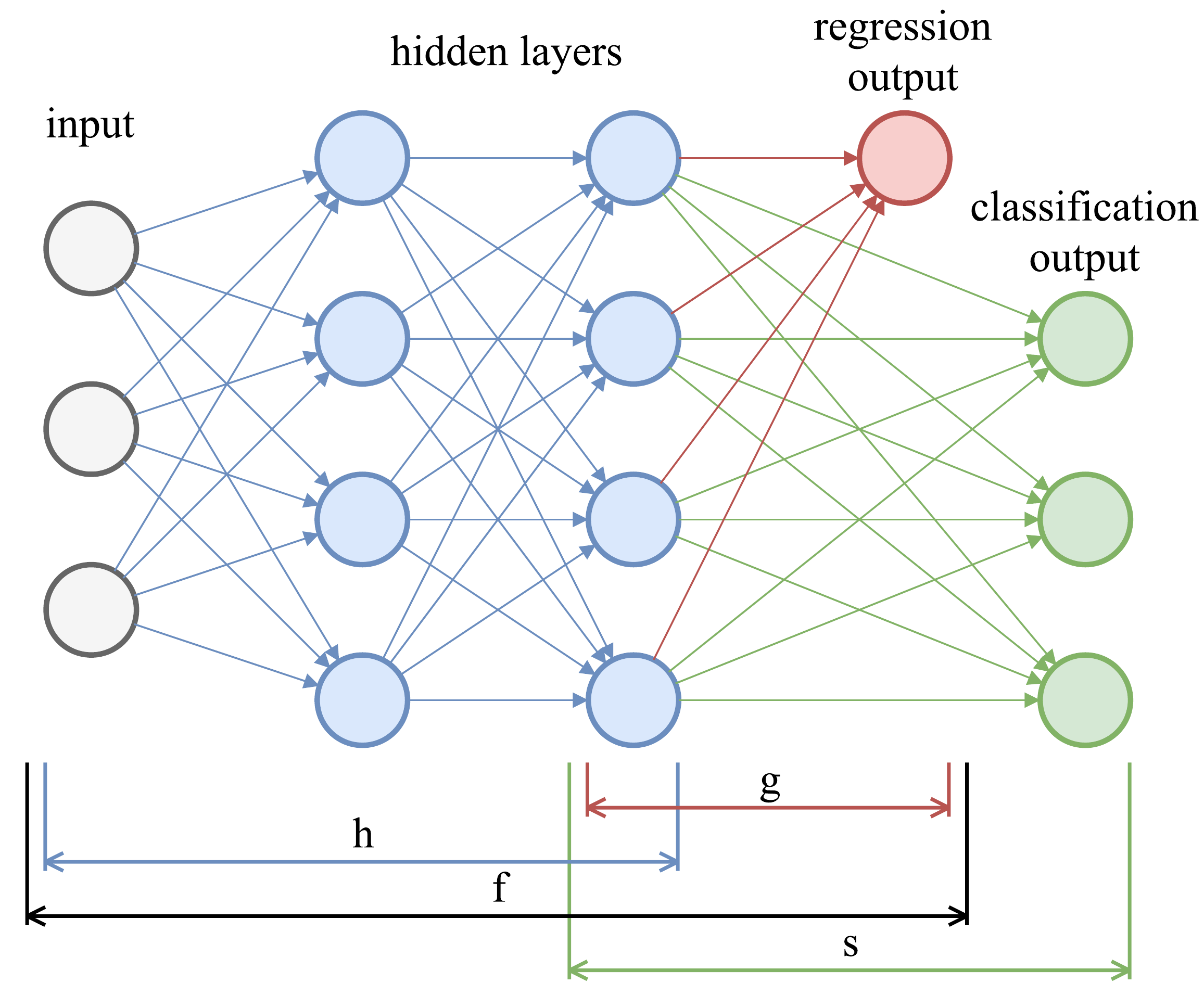}
\caption{The sample extension of the function $f(\cdot)$. The function $g(\cdot)$ always squeezes the last hidden layer into one value. On the other hand the function $s(\cdot)$ may have hidden layers, but the simplest architecture is presented.}
\label{drawering_example}
\end{figure}

Our main idea is to extend the neural network $f(\cdot)$. For every input $x$ the value of the last hidden layer $z = h(x)$ is duplicated and processed by two independent, parameterized functions. The first of them is $g(\cdot)$ as before and the second one is called $s(\cdot)$. The original neural network $g(h(\cdot))$ is trained to minimize the given loss function $\mathcal{L}_f$ and the neural network $s(h(\cdot))$ is trained with a new loss function $\mathcal{L}_s$.
An example of the extension described is presended in the Figure \ref{drawering_example}.
For the sake of consistency the loss function $\mathcal{L}_f$ will be called $\mathcal{L}_g$.

Note that functions $g(h(\cdot))$ and $s(h(\cdot))$ share parameters, because $g(h(\cdot))$ and $s(h(\cdot))$ are compositions having the same inner function. Since the parameters of $h(\cdot)$ are shared. It means that learning $s(h(\cdot))$ influences $g(h(\cdot))$ (and the other way around). We want to train all these functions jointly which may be hard in general, but the function $s(\cdot)$ and the loss function $\mathcal{L}_s$ are constructed in a special way presented below.

All real values are clustered into $n$ consecutive intervals i.e. disjoint sets $e_1, e_2, ..., e_n$ such that
\begin{itemize}
\item $\cup_{i=1}^n e_i$ covers all real numbers,
\item $r_j < r_k$ for $r_j \in e_j, r_k \in e_k$, when $j < k$.
\end{itemize}

The function $s(h(\cdot))$ (evaluated for an input $x$) is trained to predict which of the sets $(e_i)_{i=1}^n$ contains $y$ for a pair $(x,y) \in D$. The loss function $\mathcal{L}_s$ may be defined as a non-binary cross-entropy loss which is typically used in classifiation problems. In the simplest form the function $s(\cdot)$ may be just a multiplication by a matrix $S$ (whose first dimension is $n$).

To sum up, \emph{drawering} in its basic form is to add additional, parallel layer which takes as the input the value of the last hidden layer of the original neural network $f(\cdot)$. A modified (\emph{drawered}) neural network is trained to predict not only the original target, but also additional one which depicts order of magnitude of the original target. As a result extended neural network simultaneously solves the regression task and a related classification problem.

One possibility to define sets $e_i$, called \emph{drawers}, is to take suitable percentiles of target values to make each $e_i$ contain roughly the same number of them.

\subsection{Training functions $g(h(\cdot))$ and $s(h(\cdot))$ jointly} \label{howToTrainFunctions}
Training is done using gradient descent, hence it is sufficient to obtain gradients of all functions defined i.e. $h(\cdot)$, $g(\cdot)$ and $s(\cdot)$. For a given pair $(x,y) \in D$ the forward pass for $g(h(x))$ and $s(h(x))$ is calculated (note that a majority of calculations is shared). Afterwards two backpropagations are processed.

The backpropagation for the composition $g(h(x))$ using loss function $\mathcal{L}_g$ returns a vector which is a concatenation of two vectors  $grad_g$ and $grad_{h,g}$, such that $grad_g$ is the gradient of function $g(\cdot)$ at the point $h(x)$ and $grad_{h,g}$ is the gradient of function $h(\cdot)$ at the point $x$. Similarly, the backpropagation for $s(h(x))$ using loss function $\mathcal{L}_s$ gives two gradients $grad_s$ and $grad_{h,s}$ for functions $s(\cdot)$ and $h(\cdot)$, respectively.

The computed gradients of $g(\cdot)$ and $s(\cdot)$ parameters (i.e. $grad_g$ and $grad_s$) can be applied as in the normal case -- each one of those functions takes part in only one of the backpropagations.

Updating the parameters belonging to the $h(\cdot)$ part is more complex, because we are obtaining two different gradients $grad_{h,g}$ and $grad_{h,s}$. It is worth noting that $h(\cdot)$ parameters are the only common parameters of the compositions $g(h(x))$ and $s(h(x))$. We want to take an average of the gradients $grad_{h,g}$ and $grad_{h,s}$ and apply (update $h(\cdot)$ parameters). Unfortunately, the orders of magnitute of them may be different. Therefore, taking an unweighted average may result in minimalizing only one of the loss functions $\mathcal{L}_g$ or $\mathcal{L}_s$. To address this problem, the averages $a_g$ and $a_s$ of absolute values of both gradients are calculated.

Formally, the norm $L^1$ is used to define:
\begin{equation}
a_g = \left\lVert grad_{h,g} \right\rVert_1,
\end{equation}
\begin{equation*}
a_s = \left\lVert grad_{h,s} \right\rVert_1.
\end{equation*}
The values $a_g$ and $a_s$ aproximately describe the impacts of the loss functions $\mathcal{L}_g$ and $\mathcal{L}_s$, respectively. The final vector $grad_h$ which will be used as a gradient of $h(\cdot)$ parameters in the gradient descent procedure equals:
\begin{equation}
grad_h = \alpha grad_{h,g} + (1 - \alpha) \frac{a_g}{a_s} grad_{h,s}
\end{equation}
for a hyperparameter $\alpha \in (0,1)$, typically $\alpha = 0.5$. This strategy makes updates of $h(\cdot)$ parameters be of the same order of magnitude as in the procces of learning the original neural network $f(\cdot)$ (without \emph{drawering}).

One can also normalize the gradient $grad_{h,g}$ instead of the gradient $grad_{h,s}$, but it may need more adjustments in the hyperparameters of the learning procedure (e.g. learning rate alteration may be required).

Note that for $\alpha = 1$ the learning procedure will be identical as in the original case where the function $f$ is trained using loss function $\mathcal{L}_g$ only.

It is useful to bear in mind that both backpropagations also share a lot of calculations. In the extreme case when the ratio $\frac{a_c}{a_d}$ is known in advance one backpropagation may be performed simultaneously for loss function $\mathcal{L}_g$ and the weighted loss function $\mathcal{L}_s$. We noticed that the ratio needed is roughly constant between batches iterations therefore may be calculated in the initial phase of learning. Afterwards may be checked and updated from time to time.

\textit{In this section we slightly abused the notation -- a value of gradient at a given point is called just a gradient since it is obvious what point is considered.}
\subsection{Defining \emph{drawers}}
\subsubsection{Regular and uneven}\label{evenUneven}
We mentioned in the subsection \ref{firtsMentionOfPercentiles} that the simplest way of defining \emph{drawers} is to take intervals whose endings are suitable percentiles that distribute target values uniformly. In this case $n$ \emph{regular drawers} are defined in the following way:
\begin{equation}
e_i = (q_{i-1,n}, q_{i,n}]
\end{equation}
 where $q_{i,n}$ is $\frac{i}{n}$-quantile of targets $y$ from training set (the values $q_{0,n}$ and $q_{n,n}$ are defined as \emph{minus infinity} and \emph{plus inifinity}, respectively).
This way of defining \emph{drawers} makes each interval $e_i$ contain approximately the same number of target values.

However, we noticed that an alternative way of defining $e_i$'s, which tends to support classical mean square error (MSE) loss better, may be proposed. The MSE loss penalizes more when the difference between the given target and the prediction is larger. To address this problem $drawers$ may be defined in a way which encourages the learning procedure to focus on extreme values. \emph{Drawers} should group the middle values in bigger clusters while placing extreme values in smaller ones. The definition of $2n$ \emph{uneven drawers} is as follows:
\begin{equation}
e_i = (q_{1,2^{n-i+2}}, q_{2,2^{n-i+2}}], \text{ for } i \leq n,
\end{equation}
\begin{equation*}
e_i = (q_{2^{i-n+1}-2,2^{i-n+1}}, q_{2^{i-n+1}-1,2^{i-n+1}}], \text{ for } i>n.
\end{equation*}
In this case every \emph{drawer} $e_{i+1}$ contains approximately two times more target values as compared to \emph{drawer} $e_i$ for $i<n$. Finally, both $e_n$ and $e_{n+1}$ contain the maximum of $25\%$ of all target values. Similarly to the asceding intervals in the first half, $e_i$ are desceding for $i>n$ i.e. contain less and less target values.

The number of \emph{drawers} $n$ is a hyperparameter. The bigger $n$ the more complex distribution may be modeled. On the other hand each \emph{drawers} has to contain enough representants among targets from training set. In our experiments each \emph{drawer} contained at least 500 target values.

\subsubsection{Disjoint and nested} \label{disjointNested}
We observed that sometimes it may be better to train $s(h(\cdot))$ to predict whether target is in a set $f_j$, where $f_j = \cup_{i=j}^n e_i$. In this case $s(h(\cdot))$ has to answer a simpler question: \textit{"Is a target higher than a given value?"} instead of bounding the target value from both sides. Of course in this case $s(h(x))$ no longer solves a one-class classification problem, but every value of $s(h(x))$ may be assesed independently by binary cross-entropy loss.\\

Therefore, \emph{drawers} may be:
\begin{itemize}
\item \emph{regular} or \emph{uneven},
\item \emph{nested} or \emph{disjoint}.
\end{itemize}
These divisions are orthogonal. In all experiments described in this paper (the section \ref{experiments}) \emph{uneven drawers} were used.

\section{Logic behind our idea}
We believe that \emph{drawering} improves learning by providing the following properties.
\begin{itemize}
\item The extension $s(\cdot)$ gives additional expressive power to a given neural network. It is used to predict additional target, but since this target is closely related with the original one, it is believed that gained knowledge is transferred to the core of the given neural network $h(\cdot)$.
\item Since categorical distributions do not assume their shape, they can model arbitrary distribution -- they are more flexible.
\item We argue that classification loss functions provide better behaved gradients than regression ones. As a result evolution of classification neural network is more smooth during learning.
\item Additional target (even closely related) works as a regularization as typically in multitask learning \cite{thrun1996learning}.
\end{itemize}

\section{Model comparison}
Effectiveness of the method presented was established with the comparison. The original and \emph{drawered} neural network were trained on the same dataset and once trainings were completed the neural networks performances on a given test set were measured.  Since \emph{drawering} affects just a learning procedure the comparision is fair.

All learning procedures depend on random initialization hence to obtain reliable results a few learning procedures in both setups were performed. Adam \cite{adam} was chosen for stochastic optimization. 

The comparison was done on two datasets described in the following section. The results are described in the section \ref{experiments}.
\section{Data}
The method presented were tested on two datasets.

\subsection{Rossmann Store Sales}
The first dataset is public and was used during \emph{Rossmann Store Sales} competition on well-known platform \emph{kaggle.com}. The official description starts as follows:

\begin{quote}
Rossmann operates over 3,000 drug stores in 7 European countries. Currently, Rossmann store managers are tasked with predicting their daily sales for up to six weeks in advance. Store sales are influenced by many factors, including promotions, competition, school and state holidays, seasonality, and locality. With thousands of individual managers predicting sales based on their unique circumstances, the accuracy of results can be quite varied.
\end{quote}

The dataset contains mainly categorical features like information about state holidays, an indicator whether a store is running a promotion on a given day etc.

Since we needed ground truth labels, only the train part of the dataset was used (in \emph{kaggle.com} notation). We split this data into new training set, validation set and test set by time. The training set ($648k$ records) is consisted of all observations before year 2015. The validation set ($112k$ records) contain all observations from January, February, March and April 2015. Finally, the test set ($84k$ records) covers the rest of the observations from year 2015.

In our version of this task target $y$ is normalized logarithm of the turnover for a given day. Logarithm was used since the turnovers are exponentially distributed. An input $x$ is consisted of all information provided in the original dataset except for \emph{Promo2} related information. A day and a month was extracted from a given date (a year was ignored).

The biggest challenge linked with this dataset is not to overfit trained model, because dataset size is relatively small and encoding layers have to be used to cope with categorical variables. Differences between scores on train, validation and test sets were significant and seemed to grow during learning. We believe that \emph{drawering} prevents overfitting -- works as a regularization in this case.

\subsection{Conversion value task} 
This private dataset depicts conversion value task i.e. a regression problem where one wants to predict the value of the next item bought for a given customer who clicked a displayed ad.

The dataset describes states of customers at the time of impressions. The state (input $x$) is a vector of mainly continuous features like a price of last item seen, a value of the previous purchase, a number of items in the basket etc. Target $y$ is the price of the next item bought by the given user. The price is always positive since only users who clicked an ad and converted are incorporated into the dataset.

The dataset was split into training set ($2,1$ million records) and validation set ($0,9$ million observations). Initially there was also a test set extracted from validation set, but it turned out that scores on validation and test sets are almost identical.

We believe that the biggest challenge while working on the conversion value task is to tame gradients which vary a lot. That is to say, for two pairs $(x_1, y_1)$ and $(x_2, y_2)$ from the dataset, the inputs $x_1$ and $x_2$ may be close to each other or even identical, but the targets $y_1$ and $y_2$ may even not have the same order of magnitude. As a result gradients may remain relatively high even during the last phase of learning and the model may tend to predict the last encountered target ($y_1$ or $y_2$) instead of predicting an average of them. We argue that \emph{drawering} helps to find general patterns by providing better behaved gradients.

\section{Experiments}\label{experiments}
In this section the results of the comparisons described in the previous section are presented.

\subsection{Rossmann Store Sales}
In this case the original neural network $f(\cdot)$ takes an input which is produced from 14 values -- 12 categorical and 2 continuous ones. Each categorical value is encoded into a vector of size $min(k,10)$, where $k$ is the number of all possible values of the given categorical variable. The minimum is applied to avoid incorporating a redundancy. Both continuous features are normalized. The concatenation of all encoded features and two continuous variables produces the input vector $x$ of size 75.

The neural network $f(\cdot)$ has a sequential form and is defined as follows:
\begin{itemize}
\item an input is processed by $h(\cdot)$ which is as follows:
	\begin{itemize}
	\item $Linear(75, 64)$,
	\item $ReLU$,
	\item $Linear(64, 128)$,
	\item $ReLU$,
	\end{itemize}
\item afterwards an output of $h(\cdot)$ is fed to a simple function $g(\cdot)$ which is just a $Linear(128, 1)$.
\end{itemize}
The \emph{drawered} neural network with incorporated $s(\cdot)$ is as follows:
\begin{itemize}
\item as in the original $f(\cdot)$, the same $h(\cdot)$ processes an input,
\item an output of $h(\cdot)$ is duplicated and processed independently by $g(\cdot)$ which is the same as in the original $f(\cdot)$ and $s(\cdot)$ which is as follows:
	\begin{itemize}
	\item $Linear(128, 1024)$,
	\item $ReLU$,
	\item $Dropout(0.5)$,
	\item $Linear(1024, 19)$,
	\item $Sigmoid$.
	\end{itemize}
\end{itemize}
\emph{The torch notation were used, here:
\begin{itemize}
\item $Linear(a, b)$ is a linear transformation -- vector of size $a$ into vector of size $b$,
\item $ReLU$ is the rectifier function applied pointwise,
\item $Sigmoid$ ia the sigmoid function applied pointwise,
\item $Dropout$ is a dropout layer \cite{srivastava2014dropout}.
\end{itemize}
}

The \emph{drawered} neural network has roughly $150k$ more parameters. It is a huge advantage, but these additional parameters are used only to calculate new target and additional calculations may be skipped during an evaluation. We believe that patterns found to answer the additional target, which is related to the original one, were transferred to the core part $h(\cdot)$.

We used dropout only in $s(\cdot)$ since incorporating dropout to $h(\cdot)$ causes instability in learning. While work on regression tasks we noticed that it may be a general issue and it should be investigated, but it is not in the scope of this paper.

Fifty learning procedures for both the original and the extended neural network were performed. They were stopped after fifty iterations without any progress on validation set (and at least one hundred iterations in total). The iteration of the model which performed the best on validation set was chosen and evaluated on the test set. The loss function used was a classic square error loss.

The minimal error on the test set achieved by the \emph{drawered} neural network is $4.481$, which is $7.5\%$ better than the best original neural network. The difference between the average of Top5 scores is also around $7.5\%$ in favor of \emph{drawering}. While analyzing the average of all fifty models per method the difference seems to be blurred. It is caused be the fact that a few learning procedures overfited too much and achieved unsatisfying results. But even in this case the average for \emph{drawered} neural networks is about $3.8\%$ better. All these scores with standard deviations are showed in the Table \ref{rossmannScores}.

\begin{table}[!h]
\renewcommand{\arraystretch}{1.3}
\caption{Rossmann Store Sales Scores}
\label{rossmannScores}
\centering
\begin{tabular}{c||c|c|c|c|c}
Model & Min & Top5 mean & Top5 std & All mean & All std\\
\hline
Original & $4.847$ & $4.930$ & $0.113$ & $5.437$ & $0.259$\\
Extended & $4.481$ & $4.558$ & $0.095$ & $5.232$ & $0.331$\\
\end{tabular}
\end{table}


\textit{We have to note that extending $h(\cdot)$ by additional $150k$ parameters may result in ever better performance, but it would drastically slow an evaluation. However, we noticed that simple extensions of the original neural netwok $f(\cdot)$ tend to overfit and did not achieve better results.}

The train errors may be also investigated. In this case the original neural network performs better which supports our thesis that \emph{drawering} works as a regularization. Detailed results are presented in the Table \ref{rossmannScoresTrain}.

\begin{table}[!h]
\renewcommand{\arraystretch}{1.3}
\caption{Rossmann Store Sales Scores on Training Set}
\label{rossmannScoresTrain}
\centering
\begin{tabular}{c||c|c|c|c|c}
Model & Min & Top5 mean & Top5 std & All mean & All std\\
\hline
Original & $3.484$ & $3.571$ & $0.059$ & $3.494$ & $0.009$\\
Extended & $3.555$ & $3.655$ & $0.049$ & $3.561$ & $0.012$\\
\end{tabular}
\end{table}

\subsection{Conversion value task}
This dataset provides detailed users descriptions which are consisted of 6 categorical features and more than 400 continuous ones. After encoding the original neural network $f(\cdot)$ takes an input vector of size 700. The core part $h(\cdot)$ is the neural network with 3 layers that outputs a vector of size 200. The function $g(\cdot)$ and the extension $s(\cdot)$ are simple, $Linear(200,1)$ and $Linear(200, 21)$, respectively.

In case of the conversion value task we do not provide a detailed model description since the dataset is private and this experiment can not be reproduced. However, we decided to incorporate this comparison to the paper because two versions of \emph{drawers} were tested on this dataset (\emph{disjoint} and \emph{nested}). We also want to point out that we invented \emph{drawering} method while working on this dataset and afterwards decided to check the method out on public data. We were unable to achieve superior results without \emph{drawering}. Therefore, we believe that work done on this dataset (despite its privacy) should be presented.

To obtain more reliable results ten learning procedures were performed for each setup:
\begin{itemize}
\item \textit{Original} -- the original neural network $f(\cdot)$,
\item \textit{Disjoint} -- \emph{drawered} neural network for \emph{disjoint drawers},
\item \textit{Nested} -- \emph{drawered} neural network for \emph{nested drawers}.
\end{itemize}

\begin{figure}[!t]
\centering
\includegraphics[width=\columnwidth]{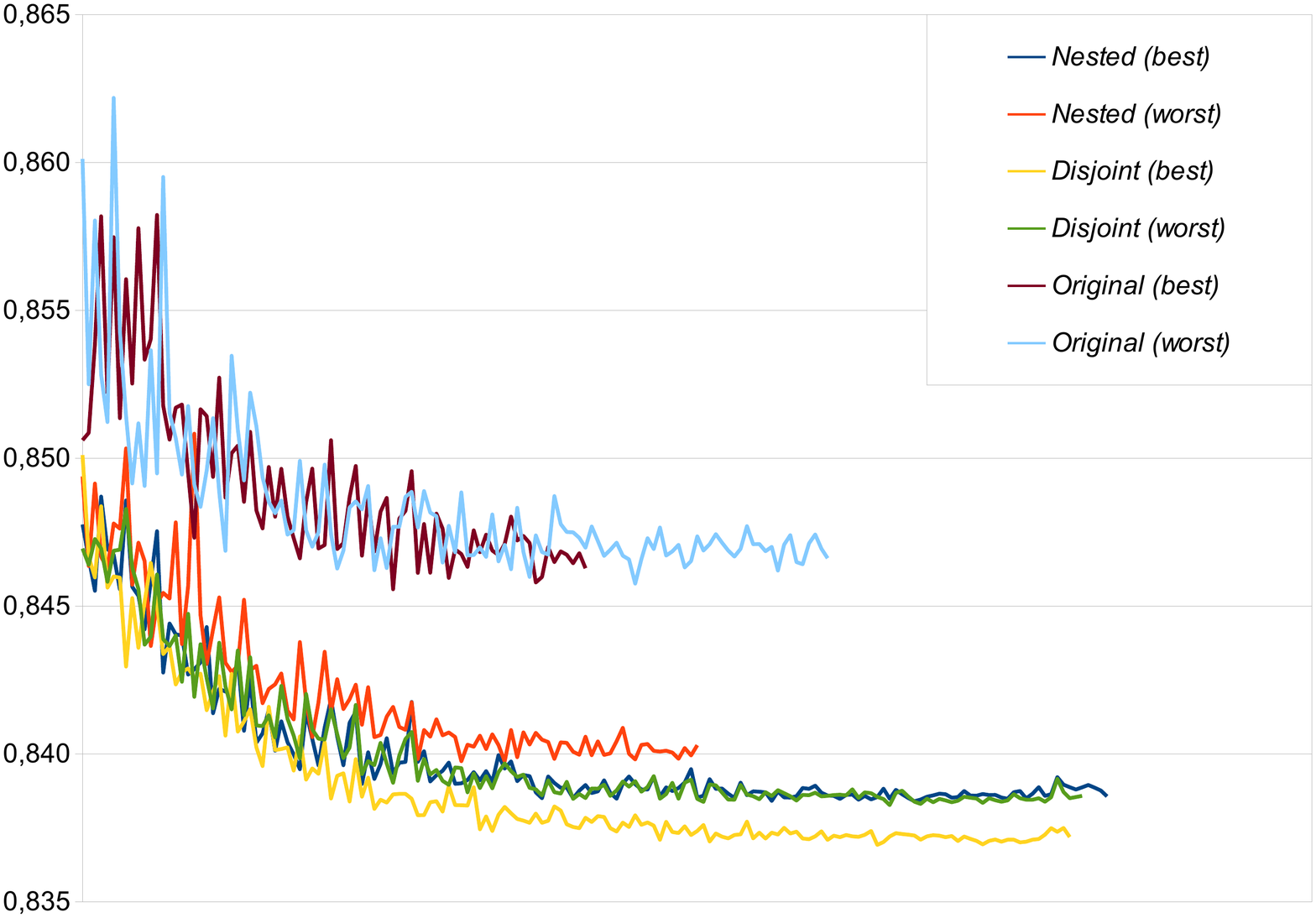}
\caption{Sample evolutions of scores on validation set during learning procedures.}
\label{cvcurves}
\end{figure}

In the Figure \ref{cvcurves} six learning curves are showed. For each of three setups the best and the worst ones were chosen. It means that the minimum of the other eight are between representants showed. The first 50 iterations were skipped to make the figure more lucid. Each learning procedure was finished after 30 iterations without any progress on the validation set.

It may be easily inferred that all twenty \emph{drawered} neural networks performed significantly better than neural networks trained without the extension. The difference between \textit{Disjoint} and  \textit{Nested} versions is also noticeable and $\textit{Disjoint}$ \emph{drawers} tends to perform slightly better.

In the Rossmann Stores Sales case we experienced the opposite, hence the version of \emph{drawers} may be understood as a hyperparameter. We suppose that it may be related with the size of a given dataset.

\section{Analysis of $s(h(x))$ values}

Values of $s(h(x))$ may be analyzed. For a pair \mbox{$(x,y) \in D$} the $i$-th value of the vector $s(h(x))$ is the probability that target $y$ belongs to the \emph{drawer} $f_i$. In this section we assume that \emph{drawers} are nested, hence values of $s(h(x))$ should be descending. Notice that we do not force this property by the architecture of \emph{drawered} neural network, so it is a side effect of the nested structure of \emph{drawers}.

\begin{figure}[!t]
\centering
\includegraphics[width=\columnwidth]{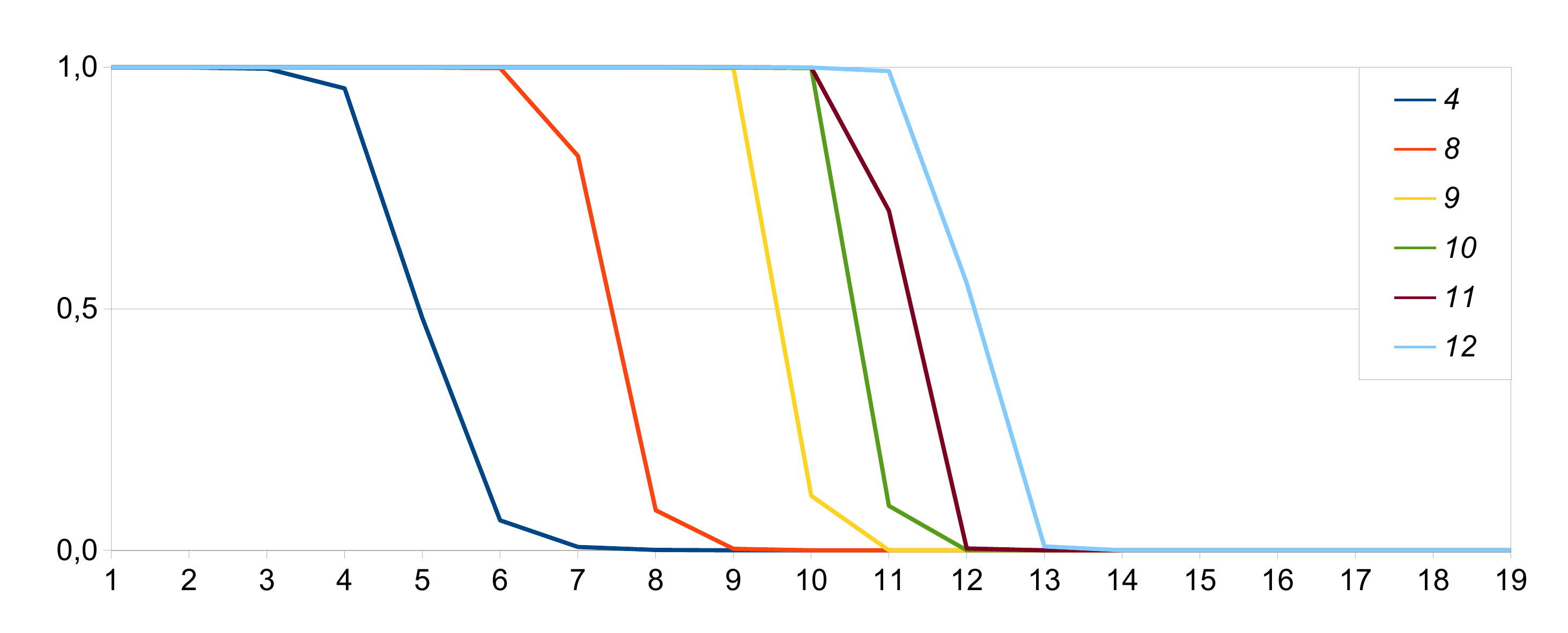}
\caption{Sample values of $s(h(x))$ for a randomly chosen model solving Rossmann Store Sales problem (nested \emph{drawers}).}
\label{rss}
\end{figure}

In the Figure \ref{rss} a few sample distributions are showed. Each according label is ground truth ($i$ such that $e_i$ contains target value). The values of $s(h(x))$ are clearly monotonous as expected. It seems that $s(h(x))$ performs well -- values are close to one in the beginning and to zero in the end. A switch is in the right place, close to the ground truth label and misses by maximum one \emph{drawer}.



\section{Conclusion}
The method presented, \emph{drawering}, extends a given regression neural network which makes training more effective. The modification affects the learning procedure only, hence once \emph{drawered} model is trained, the extension may be easily omitted during evaluation without any change in prediction. It means that the modified model may be evaluated as fast as the original one but tends to perform better.
\newpage
We believe that this improvement is possible because \emph{drawered} neural network has bigger expressive power, is provided with better behaved gradients, can model arbitrary distribution and is regularized. It turned out that the knowledge gained by the modified neural network is contained in the parameters shared with the given neural network.

Since the only cost is an increase in learning time, we believe that in cases when better performance is more important than training time, \emph{drawering} should be incorporated into a given regression neural network.



\bibliographystyle{IEEEtran}
\bibliography{szufladki}

\begin{thebibliography}{1}
\providecommand{\url}[1]{#1}
\csname url@samestyle\endcsname
\providecommand{\newblock}{\relax}
\providecommand{\bibinfo}[2]{#2}
\providecommand{\BIBentrySTDinterwordspacing}{\spaceskip=0pt\relax}
\providecommand{\BIBentryALTinterwordstretchfactor}{4}
\providecommand{\BIBentryALTinterwordspacing}{\spaceskip=\fontdimen2\font plus
\BIBentryALTinterwordstretchfactor\fontdimen3\font minus
  \fontdimen4\font\relax}
\providecommand{\BIBforeignlanguage}[2]{{%
\expandafter\ifx\csname l@#1\endcsname\relax
\typeout{** WARNING: IEEEtran.bst: No hyphenation pattern has been}%
\typeout{** loaded for the language `#1'. Using the pattern for}%
\typeout{** the default language instead.}%
\else
\language=\csname l@#1\endcsname
\fi
#2}}
\providecommand{\BIBdecl}{\relax}
\BIBdecl

\bibitem{lecun2015deep}
Y.~LeCun, Y.~Bengio, and G.~Hinton, ``Deep learning,'' \emph{Nature}, vol. 521,
  no. 7553, pp. 436--444, 2015.

\bibitem{resnet}
\BIBentryALTinterwordspacing
K.~He, X.~Zhang, S.~Ren, and J.~Sun, ``Deep residual learning for image
  recognition,'' \emph{CoRR}, vol. abs/1512.03385, 2015. [Online]. Available:
  \url{http://arxiv.org/abs/1512.03385}
\BIBentrySTDinterwordspacing

\bibitem{wavenet}
\BIBentryALTinterwordspacing
A.~van~den Oord, S.~Dieleman, H.~Zen, K.~Simonyan, O.~Vinyals, A.~Graves,
  N.~Kalchbrenner, A.~W. Senior, and K.~Kavukcuoglu, ``Wavenet: {A} generative
  model for raw audio,'' \emph{CoRR}, vol. abs/1609.03499, 2016. [Online].
  Available: \url{http://arxiv.org/abs/1609.03499}
\BIBentrySTDinterwordspacing

\bibitem{Oord16}
\BIBentryALTinterwordspacing
A.~van~den Oord, N.~Kalchbrenner, and K.~Kavukcuoglu, ``Pixel recurrent neural
  networks,'' \emph{CoRR}, vol. abs/1601.06759, 2016. [Online]. Available:
  \url{http://arxiv.org/abs/1601.06759}
\BIBentrySTDinterwordspacing

\bibitem{thrun1996learning}
S.~Thrun, ``Is learning the n-th thing any easier than learning the first?''
  \emph{Advances in neural information processing systems}, pp. 640--646, 1996.

\bibitem{adam}
\BIBentryALTinterwordspacing
D.~P. Kingma and J.~Ba, ``Adam: {A} method for stochastic optimization,''
  \emph{CoRR}, vol. abs/1412.6980, 2014. [Online]. Available:
  \url{http://arxiv.org/abs/1412.6980}
\BIBentrySTDinterwordspacing

\bibitem{srivastava2014dropout}
N.~Srivastava, G.~E. Hinton, A.~Krizhevsky, I.~Sutskever, and R.~Salakhutdinov,
  ``Dropout: a simple way to prevent neural networks from overfitting.''
  \emph{Journal of Machine Learning Research}, vol.~15, no.~1, pp. 1929--1958,
  2014.

\end{thebibliography}

\end{document}